\pdfoutput=1

\documentclass{article}

\usepackage{iclr2025_conference,times}

\iclrfinalcopy

\usepackage{xcolor}
\usepackage{times}
\usepackage{latexsym}
\usepackage{arydshln}
\usepackage{graphicx}
\usepackage{wrapfig}

\usepackage{float}

\usepackage[T1]{fontenc}

\usepackage[utf8]{inputenc}

\usepackage{microtype}
\usepackage{comment}
\usepackage{arydshln}
\usepackage{enumitem}
\usepackage{booktabs}
\usepackage{hyperref}
\usepackage{amsmath}
\usepackage{cleveref}
\usepackage{mathrsfs}
\usepackage{mdframed}
\usepackage{tcolorbox}
\usepackage{dblfloatfix}

\usepackage{microtype}
\usepackage{algorithm}
\usepackage{algpseudocode}
\usepackage{inconsolata}
\usepackage{amsfonts}
\usepackage{amssymb}
\usepackage{multicol}

\usepackage{enumitem}
\usepackage{cleveref}
\usepackage[para]{footmisc}
\usepackage{array}
\usepackage{graphicx}
\usepackage{subcaption}
\usepackage{multirow}
\usepackage{booktabs}

\usepackage[T1]{fontenc}

\usepackage[utf8]{inputenc}
\usepackage{amsmath}
\usepackage{arydshln}

\usepackage{microtype}

\usepackage{inconsolata}

\usepackage{graphicx}
\usepackage{adjustbox}  
\usepackage{array}  
\usepackage{hhline}

\title{Layer-Wise Quantization: A Pragmatic and Effective Method for Quantizing LLMs Beyond Integer Bit-Levels} 

\author{
  \textbf{Razvan-Gabriel Dumitru\textsuperscript{1,2}},
  \textbf{Vikas Yadav\textsuperscript{1}},
  \textbf{Rishabh Maheshwary\textsuperscript{1}},
  \textbf{Paul-Ioan Clotan\textsuperscript{3}},
\\
  \textbf{Sathwik Tejaswi Madhusudhan\textsuperscript{1}},
  \textbf{Mihai Surdeanu\textsuperscript{2}}
\\
  \textsuperscript{1}ServiceNow Research,
  \textsuperscript{2}University of Arizona,
  \textsuperscript{3}Università di Bologna
\\
  \small{
    \textbf{Correspondence:} \href{razvandumm@gmail.com}{razvandumm@gmail.com}
  }
}

\begin{document}
\maketitle
\begin{abstract}
We present a simple 
meta quantization approach that quantizes different layers of a large language model (LLM) at different bit levels, 
and is independent of the underlying quantization technique.
Specifically, we quantize the most  important layers to higher bit precision and less important layers to lower bits.
We propose two effective strategies to measure the importance of layers within LLMs: 
the first measures the importance of a layer based on how different its output embeddings are from the input embeddings (higher is better); the second estimates the importance of a layer using the number of layer weights that are much larger than average (smaller is better).
We show that quantizing different layers at varying bits according to our importance scores results in minimal performance drop with a far more compressed model size. Finally, we present several practical key takeaways from our variable layer-wise quantization experiments: 
(a) LLM performance under variable quantization remains close to the original model until 25--50\% of layers are moved in lower quantization using our proposed ordering but only until 5--10\% if moved using no specific ordering; 
(b) Adding layer importance to inherently dynamic quantization techniques can further improve their performance, showing that our approach is complementary to other dynamic quantization methods;
(c) Quantizing LLMs to lower bits performs substantially better than pruning unless extreme quantization (2-bit) is used; and 
(d) Layer-wise quantization to lower bits works better in the case of larger LLMs with more layers compared to smaller LLMs with fewer layers. 
Our code is publicly available at https://github.com/RazvanDu/LayerwiseQuant/.


{\renewcommand*\@m{}
\footnotetext[1]{The work was done while Razvan-Gabriel Dumitru interned at ServiceNow.}}

\end{abstract}


\section{Introduction}



Large Language Models (LLMs) have achieved remarkable performance on a variety of tasks, especially when scaled to billions of parameters~\cite{jiang2023mistral, jiang2024mixtral, llamapaper, zhang2022opt, team2023gemini}. The largest open-source models available can have an upwards of 400B parameters, such as LLaMa3-400B \cite{llamapaper}. Even small models such as LLaMa3-8B require as much as 20GB of VRAM to run on a GPU at the original precision, making them unusable for low resource settings. In such settings 
model compression techniques such as quantization are critical \cite{zhu2023survey,wan2023efficient}. 

Most prominent techniques for model compression broadly cover pruning \cite{ma2023llm}, knowledge distillation \cite{gu2023minillm}, and quantization \cite{zhu2023survey}. Pruning yields improvements in inference speed, but often results in substantial performance drop \cite{frantar2023sparsegpt, men2024shortgpt}. On the other hand, quantization has proven to be a more robust solution for model size compression with comparatively much smaller performance drops \cite{lin2024awq}. In our work, we primarily focus on memory reduction through quantization.
Further, quantization can be training specific or post-training \cite{yao2024exploring}, where trained models are quantized without requiring any further training. We focus on post-training quantization due to its practicality.


The majority of quantization techniques proposed recently \cite[inter alia]{frantar2023gptq,xiao2023smoothquant,yao2022zeroquant} focus on the quantization of all the LLM layers to a single precision bit, or quantizing only specific weights of the network~\cite{lin2024awq}. 
This has shown to be costly, their effectiveness is data dependent, and their implementations can be considered moderately challenging.
In contrast, we propose a simple meta quantization technique that {\em quantizes different layers at different bit precision depending on their importance}. 
Figure~\ref{LIM_layerranking} shows an example of our idea.
We show that our approach is simple to implement, is independent of the underlying quantization technique (e.g., we experimented with two prominent quantization techniques:  GPT-Q~\citep{frantar2023gptq} and Quanto\footnote{\url{https://github.com/huggingface/optimum-quanto}}), performs well, and provides greater flexibility to compress models in varying bit precision as per memory requirements. 

\begin{wrapfigure}{l}{0.35\textwidth} 
\vspace{-5mm} 
\centering
\includegraphics[width=0.35\textwidth]{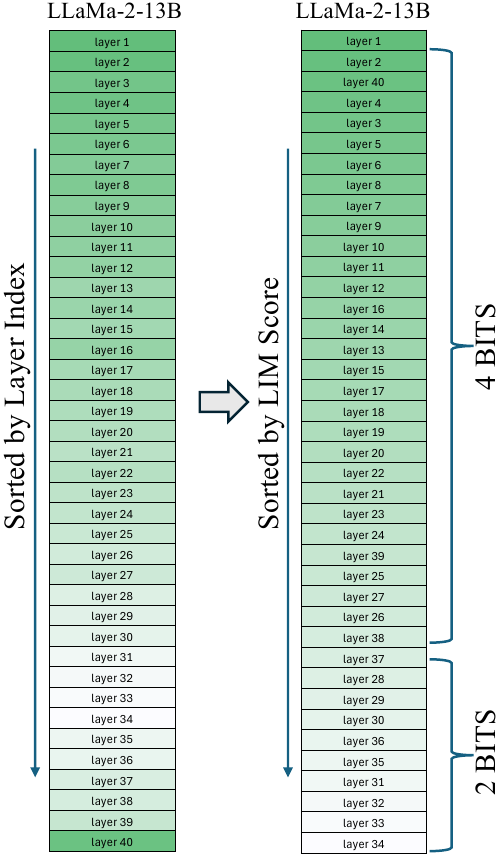} 
\caption{\small Overall intuition behind our approach. We first rank the layers in an LLM (e.g., LLaMa-2-13B here) in descending order of an importance score (shown here is ranking based on our Layer Input Modification (LIM) score, see \cref{ScoreImportance}). 
The color intensity of each layer, which represents their LIM importance score (darker color indicates higher importance score), highlights that the original layer structure (left-hand side of the figure) does {\em not} have the layers sorted according to their importance. 
This observation holds for several other LLMs (see Figure~\ref{LIM_layerrankingCommon} in the appendix). After sorting (right-hand side of the figure), the 30 most important layers are quantized in 4 bits while the remaining 10 least important layers are quantized in 2 bits, resulting in 3.5 bits as the average bit size. 
}
\label{LIM_layerranking}
\vspace{-7mm} 
\end{wrapfigure}

 
 The contributions of our work are as follows:

\begin{itemize}[itemsep=0em,topsep=0em,  wide, labelwidth=!, labelindent=0pt]
  \item We introduce a novel method for layer-wise quantization of LLMs at different bit levels to achieve flexible lower bit precision overall.
  Our method is designed to maximize performance under a given memory budget by keeping as many layers as possible in higher precision.
  We accompany our method with a detailed study, which highlights various important findings such as importance of layer ranking for quantizing less important layers with lower bits and more important layers with higher bit precision. We show that models can be quantized significantly more from 4-bit while retaining 90\% of performance when following our proposed order. 
  
  \item We propose and study two layer importance scores for variable quantization. To our knowledge, we are the first to propose layer orderings for quantization. 
  Our first score, named {\em layer input modification} (LIM), is based on how much a layer changes its input representations into the output ones. This score is calculated with an unlabeled text calibration corpus. The second scoring method, called {\em z-score distribution} (ZD), measures the distribution of parameter weights within a layer to determine its importance. 
  Thus, ZD does not require calibration data.
  We validate these scores by empirically showing that when LLM layers are quantized to lower bits as per rankings from our two importance scores, they retain performance much more strongly than several other layer ordering baselines. 
  We observe that LIM performs better than ZD on average but the differences are not large, and ZD has less performance variance for different LLMs. We consider this a success for ZD, which is simpler and does not require calibration data.

  
  
  \item We evaluate the impact of our variable quantization method based on layer importance on five top performing LLMs from different model families and different sizes. We draw several important practical findings from these experiments: 
  (a) LLM performance under variable quantization remains close to the original model when using our ordering until reaching the level of 3.0--3.25 bits on average; 
  (b) Quantizing LLMs to lower bits performs substantially better than pruning; however, under extreme quantization settings (i.e., $<=$ 2-bits) pruning shows slightly better results; (c) We show that quantization at two levels (i.e., quantizing some layers in $x$ bits and remaining layers in $y$ bits) performs much better than three levels of quantization, suggesting that the interaction between layers with different quantization is complex and may need more investigation; and (d) Layer-wise quantization to lower bits works better in the case of larger LLMs with more layers compared to smaller LLMs with fewer layers. 
  
  
\end{itemize}

\begin{figure*}[h]
\centering
\includegraphics[width=1.0\textwidth]{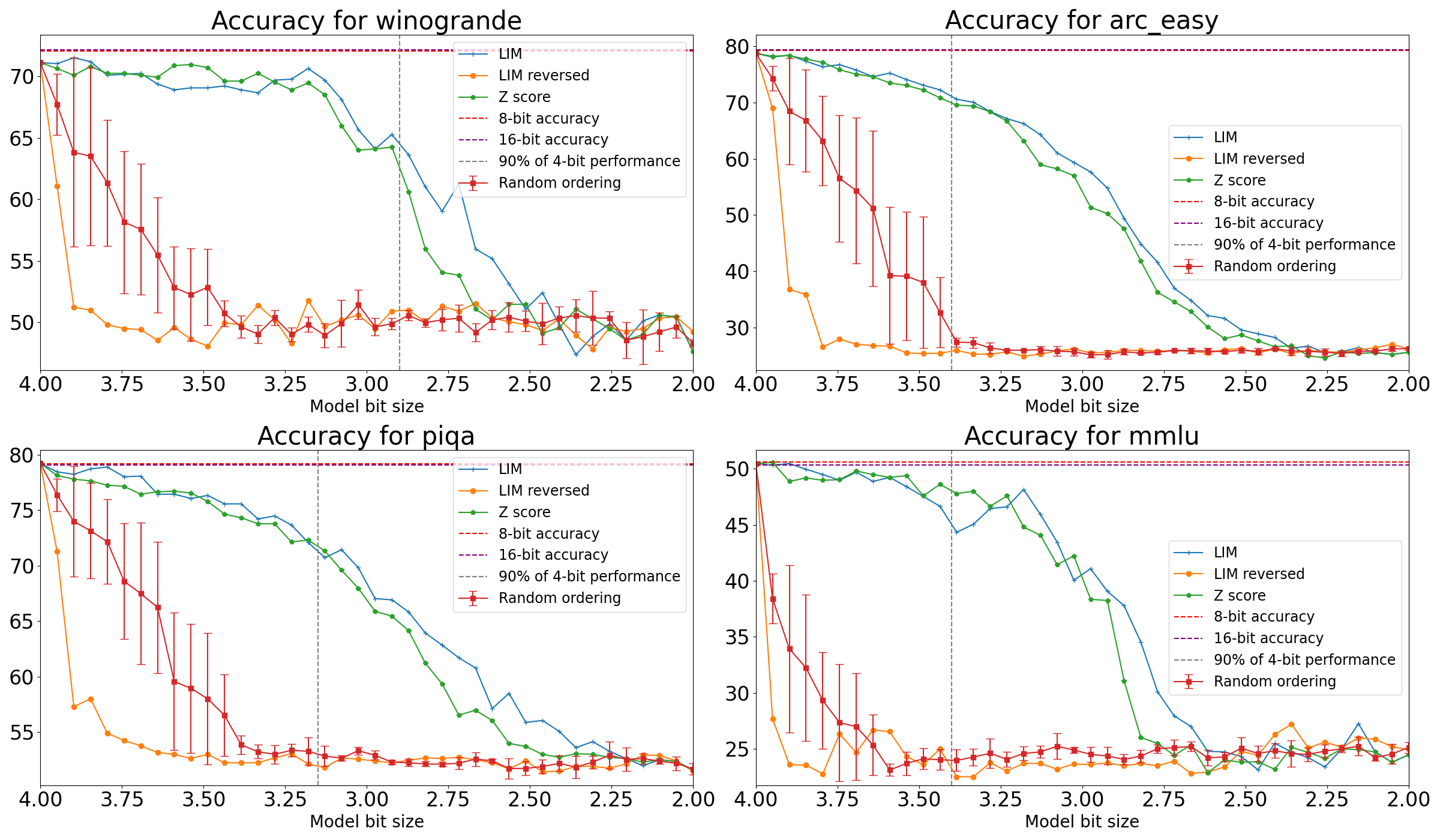}
\vspace{-7mm}
\caption{\footnotesize Plots showing the effect of variable quantization for LLaMa2-13b and multiple datasets using Quanto.
The leftmost point indicates LLM performance when all 40 layers of the LLM are represented in 4-bit; the rightmost point shows LLM performance when all layers are quantized to 2-bit. The dots on each curve (in each plot) show accuracy when the model is quantized to lower bits by converting less important layers to 2 bits one by one. 
Red and purple line indicate performance from 8bit and fp16 precision model (ceiling models). As shown, there is no considerable performance drop from fp16 or 8-bit to 4-bit precision. Hence, we focus our experiments on quantizing below 4 bits.
The vertical gray line indicates the quantization point that preserves 90\% of the 4-bit performance. The red line represents when layers are ordered randomly. We chose 3 random orders of the layers and quantized layers to 2 bits as per these orders. The standard deviation in performance from random orders are highlighted on the red curve. The curves are plotted on 2K evaluation data while results on full data is summarized in \cref{tab:42bitQuantoResults}.
The figure shows that our method retains performance much better under more aggressive quantization than all baselines.
}
\label{llama13b24bits}
\vspace{-5mm}
\end{figure*}

\begin{figure*}[h]

\centering
\includegraphics[width=1.0\textwidth]{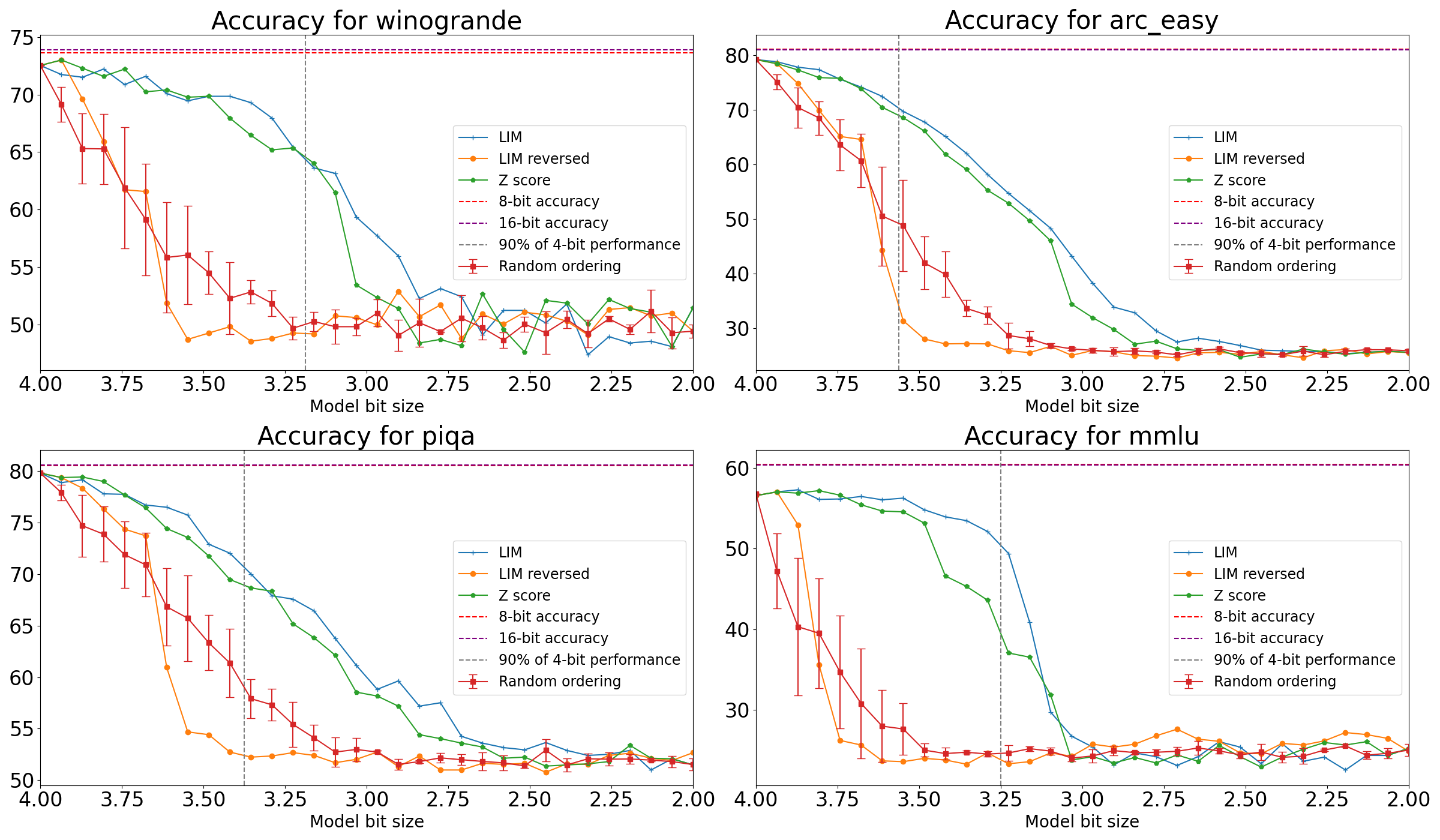}
\vspace{-7mm}
\caption{\footnotesize Plots showing the effect of variable quantization for Mistral-7b and multiple datasets using Quanto.
 All notations are the same as in \Cref{llama13b24bits}.
 Again, the figure shows that our method retains performance much better under more aggressive quantization than all baselines.}
\label{mistal24bits}
\vspace{-5mm}
\end{figure*}

\section{Related Work}



Quantization techniques for large language models (LLMs) aim to reduce the precision of weights and activations to lower-bits without significantly compromising performance~\citep{zhu2023survey}. Popular approaches include Post-Training Quantization (PTQ) \citep{banner2019post} and Quantization-Aware Training (QAT) \citep{liu2023llm}. PTQ can be divided into static quantization, which uses a small dataset to calibrate scaling factors for weights and activations, and dynamic quantization, which quantizes activations on-the-fly during inference. Our study utilizes the static PTQ techniques such as GPT-Q~\citep{frantar2023gptq} in our experiments for practicality, 
but, importantly, it is agnostic to the actual quantization technique used.

 

A few works have highlighted 4-bit precision as a robust quantization limit for wide variety of NLP tasks \citep{dettmers2023case}. Our results (\cref{llama13b24bits} and \ref{mistal24bits}) also show similar findings that performance drop is small from bf16 to 8-bits, and then to 4-bits. Hence, our experiments focus more on quantizing LLMs below 4-bits  to highlight the effectiveness of layer-wise quantization based on layer importance. Specifically, our empirical results (\cref{result_discussion}) show that variable layer-wise quantization can retain 90\% of the performance with a notable compression up to 2.85-bits overall. Concurrent (ArXiv preprint) work by \citet{tai2024mptq} also show effectiveness of variable quantization in vision language models. 

A few previous works have studied layer importance in transformers \citep{vaswani2017attention, simoulin2021many}. Some of the recent, (unpublished) concurrent works such as ShortGPT \citep{men2024shortgpt} have also proposed utilizing layer importance but primarily focusing on pruning. \citet{shen2020q} had utilized hessian information from each layer as an importance measure to quantize specific weight matrices at different bits. Different from these, our dynamic strategy focuses on utilizing layer importance for quantizing more important layers in higher bits and less important layers in lower bits. Importantly, our proposed approach of layer-wise quantization of LLMs based on their layer importance is a \textit{meta method} that can be 
coupled with any quantization techniques such as GPT-Q and Quanto (\cref{quantizationtechniquedescript}). We also compare ShortGPT's layer importance-based pruning with our layerwise quantization approach in \cref{quantvspruning}, and show that they are largely complementary.

\section{Method}




At a high level, our meta quantization strategy operates in two steps: first, given an LLM with N layers denoted as $\{\mathbf{L_1, L_2, L_3, .... L_N}\}$, we first compute the importance of each layer. Second, we then employ an existing quantization technique to quantize layers differently based on their importance.

\subsection{Layer Importance Scores}
\label{ScoreImportance}

We propose two layer importance scoring methods. 

\newcommand{\customstrut}[1]{\rule{0pt}{#1}}

We call our first layer importance score {\em layer input modification} (LIM). The intuition behind LIM is that the more a layer changes its received input embeddings the more important it must be. 
More formally, the LIM score for a specific layer \( \mathbf{L_i} \) measures the modification of the representations of \( \mathbf{L_i}\)'s input (denoted by \( \mathbf{L_i^I} \)) to create its output representations (denoted by \( \mathbf{L_i^O} \)). A near similar score has been also studied for pruning (layer removal) in a concurrent (unpublished) work \citep{men2024shortgpt}. The LIM score is the negative of cosine similarity between \( \mathbf{L_i^I}\) and \( \mathbf{L_i^O} \) as shown below:

\begin{equation}
\vspace{-2mm}
    \text{LIM}(\mathbf{L_i}) = - \frac{\mathbf{L_i^I} \cdot \mathbf{L_i^O}}{\|\mathbf{L_i^I}\| \|\mathbf{L_i^O}\|} 
     \vspace{+1mm}
\end{equation}

 The dot product \( \mathbf{L_i^I} \cdot \mathbf{L_i^O} \) quantifies the alignment between the input and output vectors, while the normalization factor \( \|\mathbf{L_i^I}\| \|\mathbf{L_i^O}\| \) scales the dot product to the range of [-1, 1], corresponding to the cosine of the angle between the two vectors. The negative sign in the LIM score indicates that a higher similarity (cosine similarity closer to 1) between \( \mathbf{L_i^I}\) and \( \mathbf{L_i^O} \) results in a lower importance score. 
For measuring change in \( \mathbf{L_i^I} \) to \( \mathbf{L_i^O} \), we use all 50 documents of pg19! \cite{rae2019compressive} (an unlabelled text corpus) as calibration corpus.  

 Our second score, called {\em z-score distribution} (ZD), is based on the distribution of parameter values and does not require any calibration data. Intuitively, this score considers a layer as more important if it contains more weights that are much higher than average.

 We examine the proportion of weights in a layer exhibiting a z-score greater than 1. The z-score of a weight \(w_i\) is defined as  
\[  
z_i = \frac{w_i - \mu}{\sigma},  
\]  
where for layer \(\mathbf{L_i}\), \(w_i\) represents an individual weight, \(\mu\) the mean of the weights, and \(\sigma\) their standard deviation. The measure of interest, \(ZD(L_{i})\), is expressed as  
\[  
\text{ZD}(L_{i}) = \frac{|Li_{Zscore} > 1|}{N_i},  
\]  
quantifying the ratio of weights whose z-scores exceed 1 (\(|Li_{Zscore} >1|\)) to the total number of weights \(N_i\) in the \(i^{th}\) layer.  

Note that both these two scores have distinct advantages and disadvantages. LIM relies on runtime information (i.e., changes in representations) so it is more aligned with inference behavior. However, LIM requires a tuning dataset to compute these representations. In contrast, ZD does not require calibration data as it uses solely the network parameters.




\paragraph{Baselines:} To highlight the benefits of our layer importance scores, we compare them to three baselines. The first baseline simply randomly ranks layers.\footnote{We use multiple different random seeds for stability.} The second baseline implements a {\em reverse} LIM ordering. Finally, to compare our approach with pruning, we implemented the pruning strategy of \citet{gromov2024unreasonable}, who has shown that removing layers from the top of the LLM to be an effective pruning strategy.

\subsection{Choosing the Number of Layers in Higher Precision Given a Memory Budget}

Under a fixed memory budget, our approach aims to keep as many layers as possible in higher precision. The maximum number of layers to keep in higher precision given available memory is easy to compute. In particular, assume we have $M_{available}$ memory available. Having all $N_{layers}$ layers in the network in lower precision requires $M_{lower}$ memory, while having all of them in higher precision requires $M_{higher}$ memory. The maximum number of layers that can be kept in higher precision is then calculated as:


\[  
N_{higher} = \lfloor \frac{M_{available}-M_{lower}}{M_{higher}-M_{lower}} * N_{layers} \rfloor
\]

For example, if we have 20GB of VRAM available; a model with 32 layers takes 17GB in lower precision (e.g., 2-bits) and 34GB in higher precision (e.g., 4-bits), then the number of layers using higher precision is: $N_{higher} = \lfloor \frac{20-17}{34-17} * 32 \rfloor = 5$. This means that 5 of the layers can be in 4-bit precision and the other $32-5=27$ have to be in 2-bit precision, for a total memory of $19.65GB<20GB$. This also underscores the practicality of our technique, as we do not add an extra hyper-parameter that requires tuning and we make use of most of the available memory.


\subsection{Quantization Techniques}
\label{quantizationtechniquedescript}
To show that our meta quantization approach is independent of the underlying quantization technique, we couple it with two well-known post-training quantization techniques: GPT-Q and Quanto.
While our focus is not on low-level quantization methods, we summarize both methods below for completeness.

\subsubsection{Quanto}

Quanto is a fast-acting quantization method\footnote{\url{https://github.com/huggingface/optimum-quanto}} that simplifies the process of reducing precision after training. In practice, Quanto achieves quicker quantization by applying uniform scaling factors across all layers of a model, avoiding the need for detailed data-driven analysis of each layer’s distribution. Similarly to most techniques, it allows for int2, int4, int8, and fp8 quantization. The Quanto library is part of Hugging Face, with symmetric per-tensor or per-channel projection for int8 and float8, and group-wise affine (with a shift or 'zero-point') for lower bit-widths, such as int2 or int4. In this paper, we used Quanto in the 2-bit and 4-bit setting, utilizing an asymmetric quantization method with RTN (round to the nearest), with the exception of Table \ref{tab:84bitQuantoResults} where we used 4-bit and 8-bit settings. In our implementation for variable quantization of different layers, we quantized the same model to two different bit levels (int2 and int4). Then, based on the layer importance order, we selected less important layers from the int2 quantized sets and more important layers from the int4 quantized sets.


\begin{table*}[ht]
    \centering   
    \footnotesize
    \begin{adjustbox}{width=\textwidth}      
    \begin{tabular}{c|c|c|c|ccccc|c}      
        \toprule 
        \multicolumn{10}{c}{{\bf Quanto Quantization}}  \\
        & Model & Avg. & Layers & \textbf{WNGD} & \textbf{ARC} & \textbf{PIQA} & \textbf{HLSWG} & \textbf{MMLU} & \textbf{Avg.Acc.} \\      
        & & bits & 2-bits & & & & & & 4-2 bits \\      
        \midrule 
        & LLaMa-7B & 4.0 & 0 & 67.9 & 74.2 & 77.3 & 55.9 & 38.4 & \textbf{62.7} \\
        & Mistral-7B & 4.0 & 0 & 72.5 & 78.8 & 79.8 & 59.3	& 55.5 & \textbf{69.2} \\
        & LLaMa-13B & 4.0 & 0 & 71.0 &	78.7 & 79.2 & 59.0	& 50.0 & \textbf{67.6} \\
        & QWEN-7B & 4.0 & 0 & 68.7 & 79.0 & 79.2 & 57.7	& 66.3 & \textbf{70.2} \\
        \cdashline{2-10}
        \customstrut{2.5ex}
        \multirow{12}{*}{\rotatebox[origin=c]{90}{{\bf LIM Ordering}}} & \multirow{3}{*}{LLaMa-7B} & 3.68 & 5 & 65.6 & 68.7 & 74.6 & 53.7 & 36.6 & \textbf{59.8} \\       
        & & 3.37 & 10 & 65.3 & 61.9 & 70.6 & 49.8 & 34.3 & \textbf{56.4} \\       
        & & 3.06 & 15 & 60.8 & 45.6 & 64.3 & 42.2 & 27.6 & 48.1 \\       
        \cline{2-10}      
        \customstrut{2.5ex}
        & \multirow{3}{*}{Mistral-7B} & 3.68 & 5 & 71.7 & 74.0 & 76.7 & 56.4 & 54.9 & \textbf{66.7} \\       
        & & 3.37 & 10 & 69.3 & 61.8 & 70.0 & 50.1 & 51.7 & 60.6 \\       
        & & 3.06 & 15 & 59.4 & 43.6 & 61.2 & 37.6 & 26.4 & 45.6 \\       
        \cline{2-10}      
        \customstrut{2.5ex}
        & \multirow{3}{*}{LLama-13B} & 3.75 & 5 & 70.2 & 76.6 & 78.0 & 57.7 & 48.9 & \textbf{66.3} \\       
        & & 3.50 & 10 & 69.1 & 72.9 & 76.3 & 55.7 & 47.4 & \textbf{64.3} \\       
        & & 3.25 & 15 & 69.7 & 66.9 & 73.7 & 52.6 & 45.8 & \textbf{61.7} \\       
        \cline{2-10}      
        \customstrut{2.5ex}
        & \multirow{3}{*}{Qwen-2-7B} & 3.64 & 5 & 51.1 & 46.3 & 67.3 & 40.9 & 24.1 & 46.0 \\       
        & & 3.28 & 10 & 51.7 & 31.0 & 57.4 & 29.8 & 23.4 & 38.6 \\       
        & & 2.92 & 15 & 48.2 & 25.7 & 53.1 & 26.1 & 24.5 & 35.5 \\       
        \midrule      
        \multirow{12}{*}{\rotatebox[origin=c]{90}{{\bf Z-score Ordering}}} & \multirow{3}{*}{LLama-7B} & 3.68 & 5 & 65.7 & 68.7 & 74.9 & 53.0 & 33.5 & \textbf{59.1} \\       
        & & 3.37 & 10 & 64.1 & 59.2 & 69.7 & 48.7 & 31.2 & 54.6 \\       
        & & 3.06 & 15 & 55.4 & 43.8 & 61.4 & 36.4 & 24.5 & 44.3 \\       
        \cline{2-10}      
        \customstrut{2.5ex}
        & \multirow{3}{*}{Mistral-7B} & 3.68 & 5 & 70.7 & 74.2 & 77.5 & 56.3 & 53.0 & \textbf{66.3} \\       
        & & 3.37 & 10 & 53.3 & 39.3 & 60.0 & 30.5 & 23.4 & 41.3 \\       
        & & 3.06 & 15 & 51.7 & 27.5 & 53.3 & 27.2 & 23.5 & 36.6 \\       
        \cline{2-10}      
        \customstrut{2.5ex}
        & \multirow{3}{*}{LLama-13B} & 3.75 & 5 & 70.3 & 76.0 & 77.2 & 57.1 & 48.1 & \textbf{65.7} \\       
        & & 3.50 & 10 & 70.7 & 72.3 & 75.8 & 54.6 & 47.0 & \textbf{64.1} \\       
        & & 3.25 & 15 & 68.9 & 66.8 & 72.1 & 51.9 & 47.0 & \textbf{61.3} \\       
        \cline{2-10}      
        \customstrut{2.5ex}
        & \multirow{3}{*}{Qwen-2-7B} & 3.64 & 5 & 63.1 & 61.0 & 70.5 & 48.4 & 55.6 & 59.7 \\       
        & & 3.28 & 10 & 51.5 & 29.3 & 53.6 & 27.0 & 25.3 & 37.3 \\       
        & & 2.92 & 15 & 49.9 & 26.0 & 52.5 & 26.0 & 25.0 & 35.9 \\       
        \bottomrule      
    \end{tabular}      
    \end{adjustbox}      
    \caption{\footnotesize Accuracy on full evaluation datasets of different models quantized with Quanto. All layers start at 4-bits; we then quantize N number of layers in 2-bits where N is mentioned in the ``Layers 2-bits'' column. We also show results for 8 to 4 bit quantization in Appendix in \cref{tab:84bitQuantoResults}.  Average performances within 90\% of the 4-bit model are highlighted in bold.}
    \label{tab:42bitQuantoResults}     
    \vspace{-4mm}
\end{table*}

\begin{table*}[ht] 
    \centering  
    \footnotesize
    \begin{tabular}{c|c|c|c|ccccc|c}  
        \toprule 
         \multicolumn{10}{c}{{\bf GPT-Q Quantization}}  \\
        & Model & Avg. & Layers & \textbf{WNGD} & \textbf{ARC} & \textbf{PIQA} & \textbf{HLSWG} & \textbf{MMLU} & \textbf{Avg.Acc} \\  
        & & bits & 2 bits & & & & & & \\
        \midrule  
        \multirow{12}{*}{\rotatebox[origin=c]{90}{{\bf LIM Ordering}}}   
        & \multirow{3}{*}{LLama-7B}   
        & 3.68 & 5 & 67.1 & 73.4 & 76.2 & 54.9 & 37.5 & 61.8 \\  
        & & 3.37 & 10 & 67.6 & 67.5 & 73.4 & 52.9 & 35.2 & 59.3 \\  
        & & 3.06 & 15 & 65.1 & 56.0 & 68.0 & 46.4 & 31.5 & 53.4 \\  
        \cline{2-10}  
        \customstrut{2.5ex}
        & \multirow{3}{*}{Mistral-7B}   
        & 3.68 & 5 & 73.1 & 77.4 & 78.0 & 59.1 & 55.4 & 68.6 \\  
        & & 3.37 & 10 & 70.6 & 74.8 & 76.9 & 56.8 & 54.9 & 66.8 \\  
        & & 3.06 & 15 & 66.1 & 65.3 & 72.9 & 50.7 & 42.0 & 59.4 \\  
        \cline{2-10}  
        \customstrut{2.5ex}
        & \multirow{3}{*}{LLama-13B}   
        & 3.75 & 5 & 71.4 & 77.5 & 78.5 & 59.0 & 49.3 & 67.1 \\  
        & & 3.50 & 10 & 71.8 & 76.4 & 77.7 & 58.2 & 48.2 & 66.5 \\  
        & & 3.25 & 15 & 72.6 & 73.7 & 75.7 & 56.7 & 47.3 & 65.2 \\  
        \cline{2-10}  
        \customstrut{2.5ex}
        & \multirow{3}{*}{Qwen-2-7B}   
        & 3.64 & 5 & 62.0 & 67.2 & 77.0 & 52.0 & 56.6 & 63.0 \\  
        & & 3.28 & 10 & 56.7 & 53.1 & 71.4 & 45.5 & 29.4 & 51.2 \\  
        & & 2.92 & 15 & 53.4 & 45.0 & 66.4 & 42.0 & 25.0 & 46.4 \\  
        \midrule  
        \multirow{12}{*}{\rotatebox[origin=c]{90}{{\bf Z-score Ordering}}}   
        & \multirow{3}{*}{LLama-7B}   
        & 3.68 & 5 & 67.2 & 71.1 & 75.9 & 54.9 & 38.1 & 61.4 \\  
        & & 3.37 & 10 & 68.0 & 66.9 & 73.1 & 52.1 & 33.7 & 58.8 \\  
        & & 3.06 & 15 & 62.9 & 56.7 & 67.8 & 46.2 & 28.5 & 52.4 \\  
        \cline{2-10}  
        \customstrut{2.5ex}
        & \multirow{3}{*}{Mistral-7B}   
        & 3.68 & 5 & 72.9 & 77.7 & 78.8 & 59.2 & 55.1 & 68.7 \\  
        & & 3.37 & 10 & 69.1 & 72.1 & 74.8 & 53.2 & 38.9 & 61.6 \\  
        & & 3.06 & 15 & 62.0 & 52.7 & 65.6 & 39.5 & 25.1 & 49.0 \\  
        \cline{2-10}  
        \customstrut{2.5ex}
        & \multirow{3}{*}{LLama-13B}   
        & 3.75 & 5 & - & - & - & - & - & - \\  
        & & 3.50 & 10 & 71.3 & 75.0 & 76.4 & 57.4 & 48.2 & 65.6 \\  
        & & 3.25 & 15 & 72.3 & 73.5 & 76.2 & 56.4 & 45.7 & 64.8 \\  
        \cline{2-10}  
        \customstrut{2.5ex}
        & \multirow{3}{*}{Qwen-2-7B}   
        & 3.64 & 5 & 69.1 & 71.1 & 75.2 & 54.0 & 64.3 & 66.8 \\  
        & & 3.28 & 10 & 65.6 & 65.2 & 71.6 & 48.2 & 47.5 & 59.6 \\  
        & & 2.92 & 15 & 54.8 & 50.0 & 66.7 & 42.8 & 30.4 & 49.4 \\  
        \bottomrule  
    \end{tabular}  
    \caption{Comparison of 4 models and their performance across various tasks with GPT-Q quantization.}  
     \label{tab:42bitGPTQResults}  
     \vspace{-2mm}
\end{table*}

\subsubsection{GPT-Q}
GPT-Q is a post-training quantization technique specifically designed for GPT models~\citep{frantar2023gptq}. Utilizing a dataset, it calculates the necessary scaling factors for quantization. After training, GPT-Q assesses the distribution of weights using this dataset to determine optimal scaling factors for converting floating-point representations to lower-bit formats such as int8 or int4. 
We only used a few data points for GPT-Q as 
it adds significant execution time overhead to our experiments. 
We modified the GPT-Q implementation in the Hugging Face library~\citep{wolf2019huggingface} for our use-case in the same way as described above for Quanto.


\section{Experiments}

\subsection{Models and Hyperparameters}

We studied quantization on 5 LLMs from different model families and different sizes -- LLaMa-2-7b \citep{llamapaper}, LLaMa-2-13B, Mistral-7b \citep{jiang2023mistral}, QWEN-1.8b, and QWEN-7b \citep{qwen}. We evaluated these LLMs on 8 A100 GPUs with a batch size of 1 for inference using an LLM harness library \citep{eval-harness}. 

\subsection{Evaluation Datasets}

We select five diverse NLP tasks for evaluating the quantization effects: Winogrande \citep{sakaguchi2021winogrande}, ARC-easy \citep{clark2018think}, PIQA \citep{bisk2020piqa}, HellaSwag \citep{zellers2019hellaswag}, and MMLU \citep{mmludataset}. We also evaluate our approach on two generation datasets to cover diverse tasks of reasoning and answer generation: GSM8K \citep{cobbe2021gsm8k}, which contains math questions, and the Natural Questions (open) dataset \citep{kwiatkowski2019natural}, which consists of open-domain answer generation task. To calibrate the LIM score we used 50 samples from PG19 \citep{raecompressive2019}, a data-set that is comprised of books published before 1919.


\section{Discussion of Results}
\label{result_discussion}


Our main results are shown in \Cref{llama13b24bits} and \ref{mistal24bits} (please also see \Cref{llama7b24bits}, \ref{qwen7b24bits}, and \ref{qwen18b24bits} in Appendix). 
In both these figures, we ranked the layers in the respective LLMs in descending order of their importance score. 
The left most point in the plots indicates that all layers of Mistral-7b and LLaMa2-7B are quantized in 4-bits; as we move to the right on x-axis, we quantize the next least important layer to 2-bits. For example, the overall bit size of 3.75 mentioned on the x-axis represents 28 (most important) layers in 4-bit and 4 (least important) layers in 2-bit. The horizontal red dotted lines indicate the performance of the entire model represented in 8-bit precision; as shown, this performs very similarly to the full model in 4-bit precision (the top left most point in \Cref{llama13b24bits} and \ref{mistal24bits}). Because the gap between the full model in 8-bit vs. 4-bit precision was less than 1\% across the majority of the datasets, we focus mostly on quantizing below 4-bits precision in our experiments.

\begin{table}[h]  
    \centering  
    \footnotesize  
    \begin{adjustbox}{width=0.5\textwidth}  
        \begin{tabular}{c|c|c|c|c|c|c}  
            \toprule  
            & \multicolumn{2}{c|}{} & \multicolumn{2}{c|}{4 to 2 bits} & \multicolumn{2}{c}{8 to 4 bits} \\  
            & Models & Layers & \textbf{GSM8K} & \textbf{NQ\_open} & \textbf{GSM8K} & \textbf{NQ\_open} \\  
            & & 2-bits & & & F1 & F1 \\
            \midrule  
            \multirow{6}{*}{\rotatebox[origin=c]{90}{{\bf LIM Ordering}}}  
            & \multirow{3}{*}{LLama-Ins}  
            & 5 & 7.5 & 15.0 & 10.5 & 36.6 \\  
            & & 10 & 1.5 & 6.7 & 13.5 & 37.9 \\  
            & & 15 & 0.5 & 3.1 & 11.0 & 35.4 \\  
            \cline{2-7}  
            \customstrut{2.5ex}
            & \multirow{3}{*}{Mistral-Ins}  
            & 5 & 24.5 & 16.2 & 34.5 & 29.6 \\  
            & & 10 & 20.0 & 9.2 & 36.5 & 29.8 \\  
            & & 15 & 4.5 & 5.1 & 34.0 & 27.7 \\  
            \midrule  
            \multirow{6}{*}{\rotatebox[origin=c]{90}{{\bf Z Ordering}}}  
            & \multirow{3}{*}{LLama-Ins}  
            & 5 & 6.0 & 11.7 & 12.5 & 37.3 \\  
            & & 10 & 3.5 & 5.3 & 12.5 & 36.9 \\  
            & & 15 & 1.0 & 1.8 & 10.0 & 34.7 \\  
            \cline{2-7}  
            \customstrut{2.5ex}
            & \multirow{3}{*}{Mistral-Ins}  
            & 5 & 25.0 & 15.4 & 37.0 & 28.5 \\  
            & & 10 & 10.5 & 8.7 & 34.5 & 30.1 \\  
            & & 15 & 1.0 & 2.2 & 34.0 & 29.3 \\  
            \bottomrule  
        \end{tabular}  
    \end{adjustbox}  
    \caption{\footnotesize Performance comparison of LLaMa-instruct-7b and Mistral-instruct-7b across two generation tasks - GSM8K and Natural Questions open split.}  
    \label{tab:generationresults}  
    \vspace{-4mm}
\end{table}

\begin{table}[htbp]
    \centering
    \begin{tabular}{lccccc}
        \toprule
        \textbf{Method} & \textbf{Layers / Threshold} & \textbf{Avg. Bits} & \textbf{Wikitext Perplexity} & \textbf{C4 Perplexity} \\
        \midrule
        \hspace{4mm}\multirow{7}{*}{    
  \raisebox{0mm}[0mm][0mm]{\rotatebox[origin=c]{90}{\textbf{Our Method}}}}
            & 32      & 3.94 & \textit{5.6134} & \textit{7.5941} \\
            & 27      & 3.90 & 5.6439 & \textbf{7.6101} \\
            & 22      & 3.85 & \textbf{5.6597} & \textbf{7.6326} \\
            & 17      & 3.81 & \textbf{5.6631} & \textbf{7.6487} \\
            & 12      & 3.76 & \textbf{5.6848} & \textbf{7.6737} \\
            & 6       & 3.71 & \textbf{5.7009} & \textbf{7.6938} \\
            & 0       & 3.65 & \textit{5.7145} & \textit{7.7126} \\
        \midrule
\hspace{4mm}\multirow{7}{*}{    
  \raisebox{0mm}[0mm][0mm]{\rotatebox[origin=c]{90}{\textbf{SpQR}}}}
            & 1.00\%  & 3.94 & \textit{5.6134} & \textit{7.5941} \\
            & 0.85\%  & 3.90 & \textbf{5.6403} & 7.6523 \\
            & 0.70\%  & 3.85 & 5.6929 & 7.6878 \\
            & 0.55\%  & 3.80 & 5.6883 & 7.7041 \\
            & 0.40\%  & 3.75 & 5.7054 & 7.7060 \\
            & 0.25\%  & 3.71 & 5.7151 & 7.7140 \\
            & 0.10\%  & 3.65 & \textit{5.7145} & \textit{7.7126} \\
        \bottomrule
    \end{tabular}
    \caption{Comparison of our method and SpQR \citep{dettmers2023spqr} on LLaMa2-7B on the same avg. bit level/memory requirement. Our technique aims to optimize performance given memory constraints, allowing quantization to fit more precisely within available memory, which, as shown, improves results over fixed bit-width quantization. For SpQR, we fixed the bit levels to 3 and beta values to 4, varying the outlier threshold from 0.1\% to 1\%. For our method, we applied SpQR with thresholds of 0.1\% to less important layers and 1\% to more important layers based on LIM ordering. The second column indicates the number of layers with a higher threshold for our method and the outlier threshold for SpQR. We bold the higher values and we use italics to denote identical values. Note that each row in the ``Our Method'' block should be compared with the row at the same position in the ``SpQR'' block. That is, our setting with 32 layers in higher precision is comparable to SpQR with the 1.00\% outlier threshold; our setting with 0 layers in higher precision is comparable with SpQR with all layers using 0.10\% outlier threshold.
    \label{tab:comparisonSPQR}
}
\end{table}

We draw the following observations from our experiments:

\begin{enumerate}[label={\bf(\arabic*)},itemsep=0em,topsep=0em,  wide, labelwidth=!, labelindent=0pt]
\item {\bf Variable quantization is useful:} The first key finding of our work is that a fixed quantization technique can be extended to a variable number of bits by quantizing different layers at different bits according to their importance. This allows LLMs 
to retain more of the original performance while fitting in a reduced memory budget.
Overall, our method of layer-wise quantization, guided by layer importance, proves to be an efficient strategy for attaining adaptable precision bit levels. 

\item {\bf Layer importance scoring is crucial:} In the figures we compare layer ranking using our LIM and ZD importance scores with ranking using a reverse LIM and random ordering. 
As seen in \cref{llama13b24bits} and \ref{mistal24bits} the quantization of least important layers from 4-bit to 2-bit as per the LIM score ranking shows strong performance retention. In contrast, quantizing based on the reverse of LIM score shows much worse performance when most important layers are quantized to 2-bit,  highlighting the strength of meaningful layer ranking. Further, LIM and ZD ranking performs substantially better than random ordering of layers baseline where we quantize layers to lower bits randomly. 
Lastly, LIM performs better than ZD on average, but the differences are not large. In some cases, i.e., Figure~\ref{qwen7b24bits}, ZD performs considerably better. We conjecture this is due to the fact that for this LLM the information gathered from the dataset used to calibrate LIM transferred less well to the evaluation datasets.
All in all, we consider this a success for ZD, which is simpler and does not require calibration data. 

\item {\bf Improving dynamic quantization techniques:} Table \ref{tab:comparisonSPQR} shows that we can use our technique to further improve quantization techniques that are inherently dynamic. Although this is not the main goal of the paper, it further underscores its usefulness and flexibility. The reason why this works is because the SOTA quantization technique SpQR \citep{dettmers2023spqr}, similarly to most dynamic techniques, choose a fixed percentage of values that will be quantized to higher precision in each layer, while our techniques adds the fact that not all layers are equally important on top of it. This also further shows that our proposed technique can be used an enhancement to any category of quantization techniques.

\item {\bf Layer-wise quantization is useful until 3.0--3.25-bits:} As shown in \cref{llama13b24bits} and \ref{mistal24bits}, our first key observation is that quantization to 8-bits barely affects performance (red vs. purple line in each plot). While there is marginal drop in performance when models are quantized to 4-bits, we observed really noticeable drops only after  the models are dropped below 3.0--3.25 bits on average using Quanto. For example, the bit size for which performance drops below 90\% on Winogrande for Mistral-7b, LLaMa-7b, QWEN-7b, and LLaMa2-13b are 3.2, 3.1, 3.85, and 2.85 respectively.

\begin{figure}[htb]
 
    \centering
    \begin{subfigure}[b]{0.47\textwidth}
    \caption{Quantizing from 8 to 4 bits against pruning}
        \includegraphics[width=\textwidth]{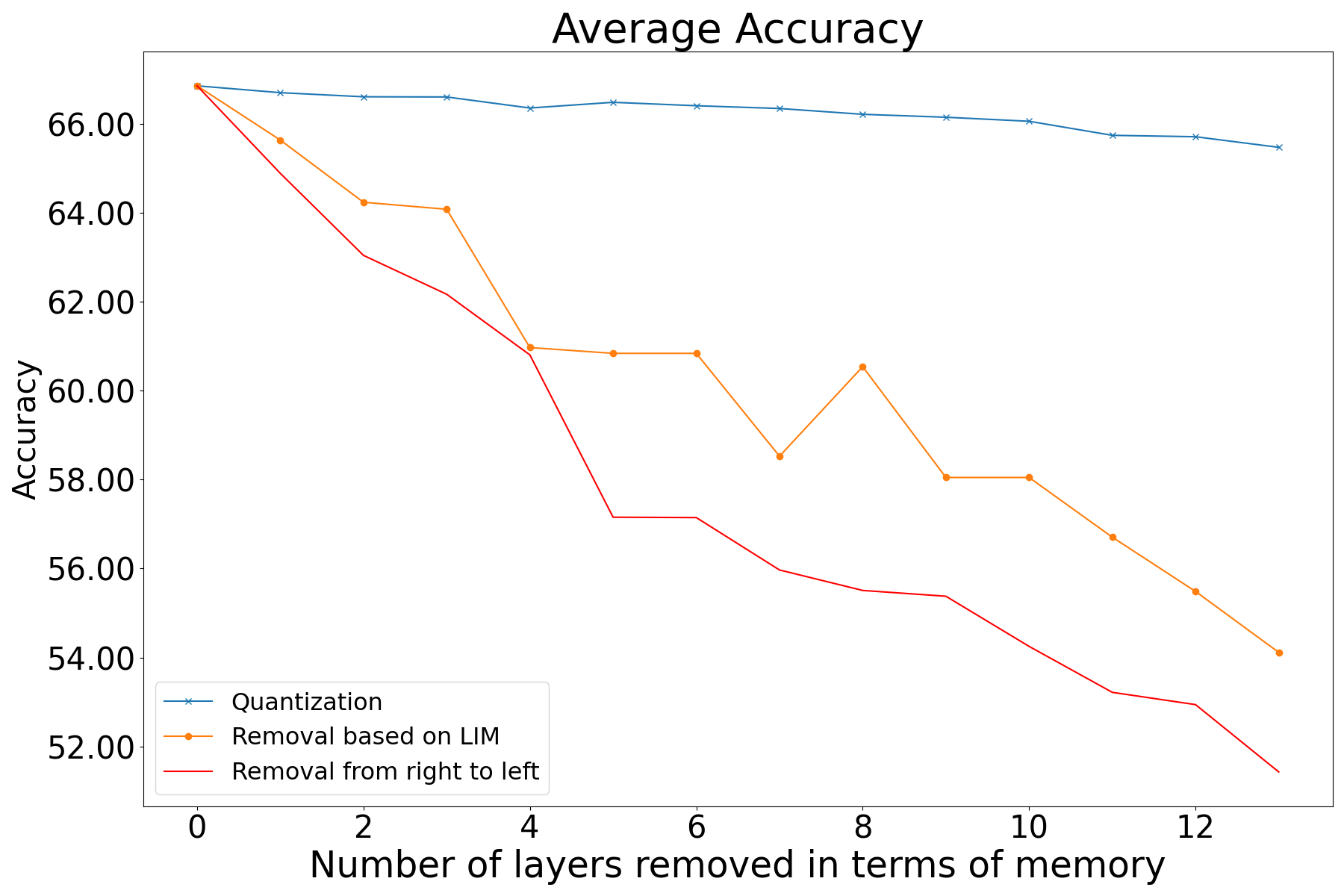}
        \label{fig:removal_8bits}
    \end{subfigure}
    \hfill 
    \begin{subfigure}[b]{0.47\textwidth}
    \caption{Quantizing from 4 to 2 bits against pruning}
        \includegraphics[width=\textwidth]{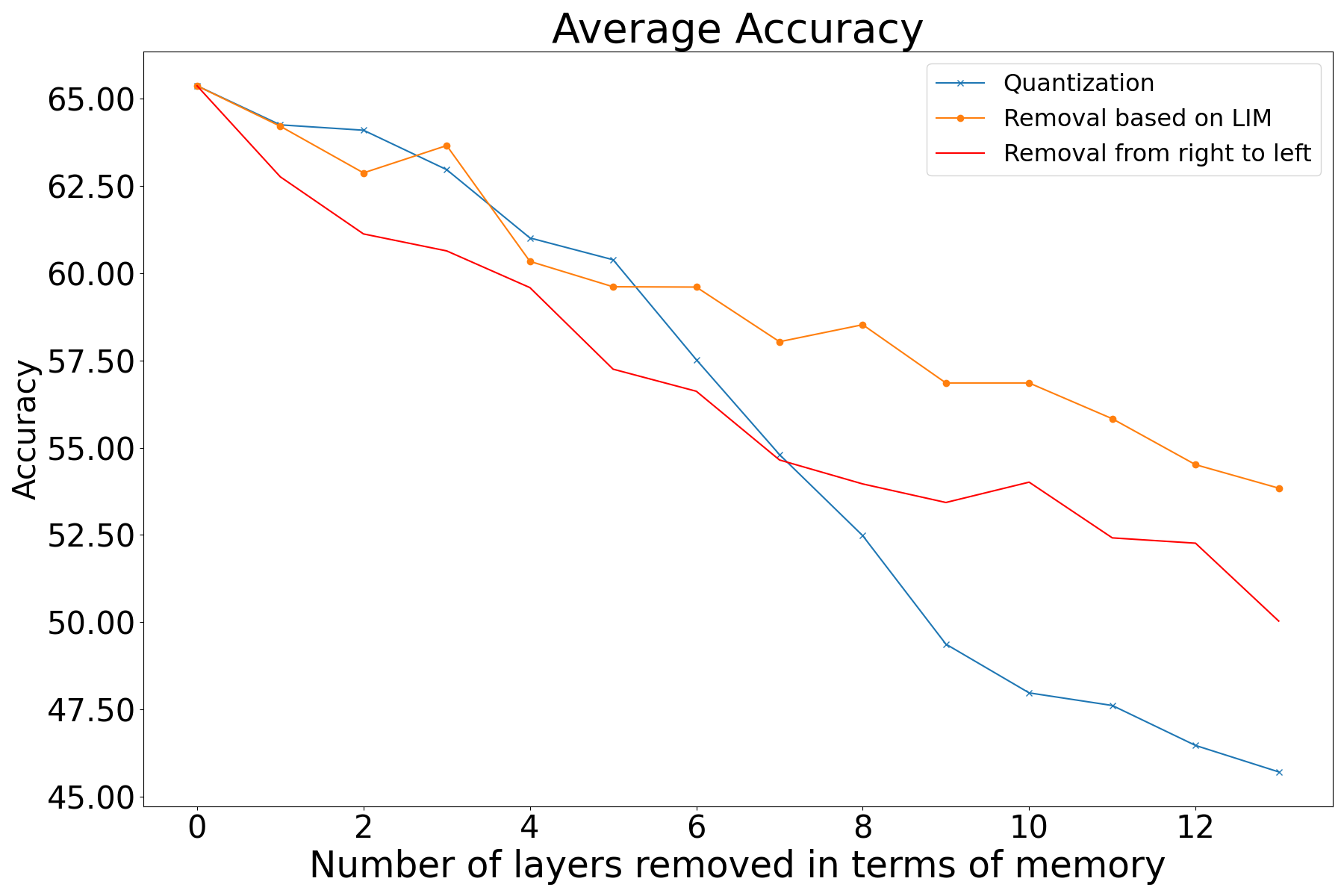}
        \label{fig:removal_4bits}
    \end{subfigure}
    \caption{We compare quantization against pruning as a method to reduce the memory requirement of a model (LLaMa-2-7b here). One increment means two layers moved to lower quantization for the blue line (quantization), and one layer removed for the red and orange lines (pruning), thus reducing the same amount of memory. We show the average accuracy over MMLU, Winogrande, PIQA, and Hellaswag.}
    \label{fig:quantization_comparisons}
    \vspace{-2mm}
\end{figure}

\item {\bf Quantization is more useful for larger LLMs:} Variable quantization of larger models (\cref{llama13b24bits}) using our importance score shows much better retention of performance with lower quantization bit precision compared to moderately-sized LLMs such as LLaMa-7B (\cref{llama7b24bits} in appendix) and Mistral-7B (\cref{mistal24bits}). Further, as we go down to even smaller LLMs such as QWEN-1.8B (see \cref{qwen18b24bits} in Appendix) with only 20 layers, we observed layer importance ranking to be not as effective. This observation aligns with many other previous works that have shown quantization to be substantially more effective for larger LLMs when compared to their smaller counterparts \cite{jin2024comprehensive}.

\item  {\bf Our method is applicable to different quantization techniques:} Tables~\ref{tab:42bitQuantoResults} and~\ref{tab:42bitGPTQResults} summarize the overall results when quantizing individual layers with Quanto and GPT-Q, respectively. 
The tables show that our method can be coupled with any other quantization techniques. On average, GPT-Q leads to an average of 4\% better accuracy than Quanto across all 5 tasks. 
Additionally, GPT-Q enables models such as LLaMa2-13B to be quantized down from 4-bits to 3.25-bits with less than a 3\% loss in average accuracy, as seen in Table \ref{tab:42bitGPTQResults}.


\item {\bf Effect on generation tasks:} We also evaluate our approach of variable quantization of different layers on generation tasks. As shown in \cref{tab:generationresults}, we observe substantial drop in performance on both GSM8K and NQ\_open generation tasks when quantizing more layers in 2-bits. Importantly, the performance drop in these generation tasks is more drastic when compared to the average performance drop in classification tasks (\cref{tab:42bitQuantoResults}), emphasizing the need for more dedicated research in quantization for generation tasks.

\end{enumerate}

\section{Analyses}

We present several analyses to further spotlight on benefits from variable layer-wise quantization.

\begin{wrapfigure}{r}{0.5\textwidth} 
\vspace{-5mm} 
\centering
\includegraphics[width=0.5\textwidth]{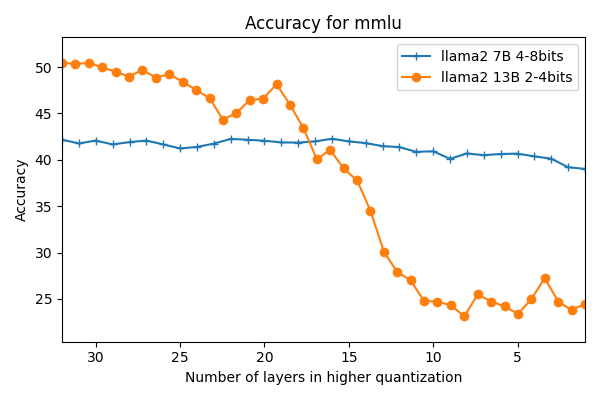} 
\caption{\footnotesize Comparison of LLaMa2-7b quantized between 8 and 4 bits with LLaMa2-13b quantized between 4 and 2 bits to check when the performance intersects.}
\label{LIM_layerranking}
\vspace{-3mm} 
\end{wrapfigure}

\subsection{Pruning vs. Quantization}
\label{quantvspruning}

We compare our variable quantization against variable pruning (using the same layer importance ranking) as an alternative for the same goal of reducing model memory requirement. In \Cref{fig:removal_8bits}, we show that for less extreme quantization levels, it is significantly better to move layers into lower quantization levels instead of removing them. For example, when 2 least important layers are removed resulting in remaining 30 layers (each in 8-bit) of LLaMa2-7b, the average performance drops to 62.7\% (shown by red line denoting work by \citet{gromov2024unreasonable}). But on the quantization counterpart with same memory i.e., when 4 layers are quantized to 4-bit and remaining 28 layers are in 8-bit (shown by blue line),  performance remains intact close to 66.8\% as shown in \cref{fig:removal_8bits}. When 12 layers are removed, the performance drops around 53\% on average while having 8 layers in 8-bits and 24 layers quantized to 4-bits (shown by blue curve) to maintain the same size, average performance still remains intact and close to 66\%. This highlights the important finding that quantization until 4-bits overall is a substantially more effective strategy compared to pruning for model compression.

On the other hand, in case of extreme levels of quantization (i.e., $< 4-$bits) as shown 
\cref{fig:removal_4bits}, it is better to plainly remove layers. For example, as shown in \cref{fig:removal_4bits} where the model is initially quantized with 
4-bits, removing $>7$ layers results in better average performance when compared to quantizing $>14$ layers in 2-bits. Thus, when model compression is requires to be the equivalent of $<3$-bits, pruning maybe the more effective strategy.  




\subsection{Quantizing Larger vs. Smaller LLMs}

We further evaluate and compare feasibility of quantizing larger LLMs more drastically (i.e., $<4$-bits by quantizing less important layers in 2-bits and keeping more important in 4-bits) or quantizing smaller LLMs moderately (i.e., $<8$-bits by quantizing less important layers in 4-bits and keeping more important in 8-bits). As shown in \cref{LIM_layerranking}, we quantize LLaMa2-13B $<4$-bits and LLaMa2-7B $<8$-bits and compare them across different memory sizes. Our empirical finding suggests it is more beneficial to quantize larger LLMs to smaller bits but only until a certain point, after which layer-wise quantizing smaller LLMs moderately ($\approx6$-bits) shows better performance.


\section{Conclusion}

We introduced a simple, flexible quantization approach that quantizes different layers at different bits based on their importance. We presented two layer importance scoring techniques which when used to select more important layers for quantizing them in 4 bits and less important layers in 2-bits lead to strong performance retention across several LLMs even until 2.85 overall bit size. Our work presents several key practical findings such as layer-wise quantization is more effective for larger LLMs (with more number of layers), in the same memory setting; quantization is better than pruning until a certain bit precision level, text generation tasks are affected more with quantization, etc. Overall, our work introduces layer-wise quantization and presents detailed empirical findings for motivating future research in this direction.



\bibliography{iclr2025_conference}
\bibliographystyle{iclr2025_conference}

\appendix

\section{Appendix}

\subsection{Quantizing Layers Using 3 Levels}
\label{threelevelquant}
\begin{figure}[h]
\centering
\includegraphics[width=0.6\textwidth]{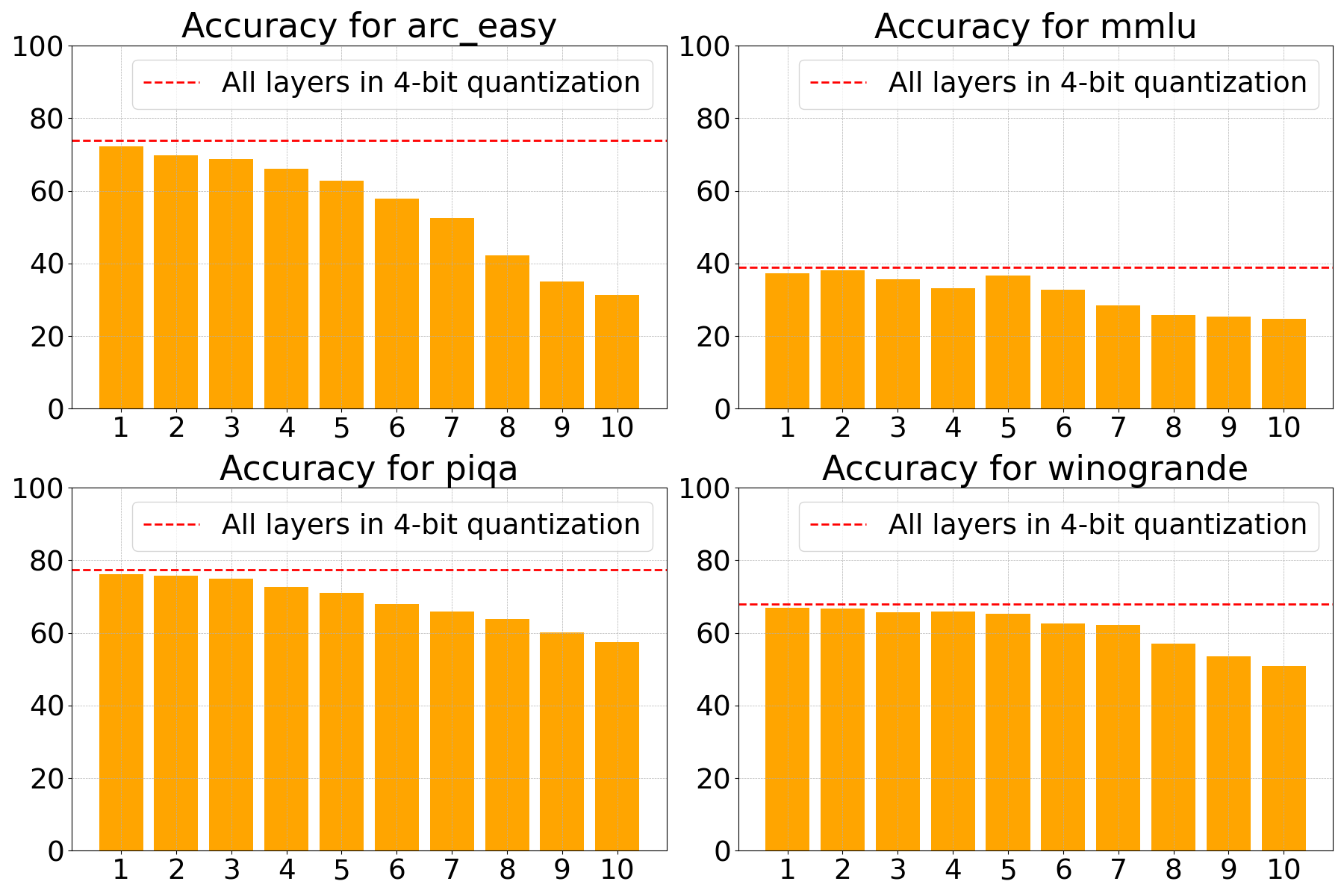}
\caption{We compare different ways to achieve 4-bit quantization using three quantization levels. Each bar going from left to right represents adding one important layer in 8 bits and moving two less important layers to 2 bits, thus keeping an average of 4-bit quantization for all of the bars. Each bar having a value x on the x-axis represents the most important x layers in 8-bits, the least important 2*x in 2-bits and the rest in 4-bits.}
\label{3levelquantization}
\end{figure}

In all of our experiments, we have focused on two level quantization, i.e., either less important layers are quantized in 2-bits and more important layers in 4-bits or less important in 4-bits and more important in 8-bits. As a plausible variant, LLM layers can also be easily quantized using three levels, i.e., least important layers in 2-bits, moderately important in 4 bits, and the most important ones in 8-bits. 
In our study, we observed three level quantization almost always performs worse than two level quantization. We show three level quantization of LLaMa2-7b with fixed overall model bit size of 4-bits. We first quantize all the 32 layers in 4-bits as shown by the leftmost bar of \cref{threelevelquant}. We then convert two least important layers to 2-bits each and one most important layer to 8-bits, thus maintaining the overall bit size of the model to 4-bits. This is represented by second bar from the left in \cref{3levelquantization}. We repeat the same process of converting two more layers in 2-bits and a more important layer to 8-bits represented by the consecutive bars in \cref{3levelquantization}. As observed, three level quantization always performs worse than one level quantization when the target bit-level is the same, thus we don't propose the technique as a way to achieve better performance for a set quantization level, but as a way to achieve a variable level of quantization while retaining maximum performance.

\begin{figure}[t]
\centering
\includegraphics[width=0.55\textwidth]{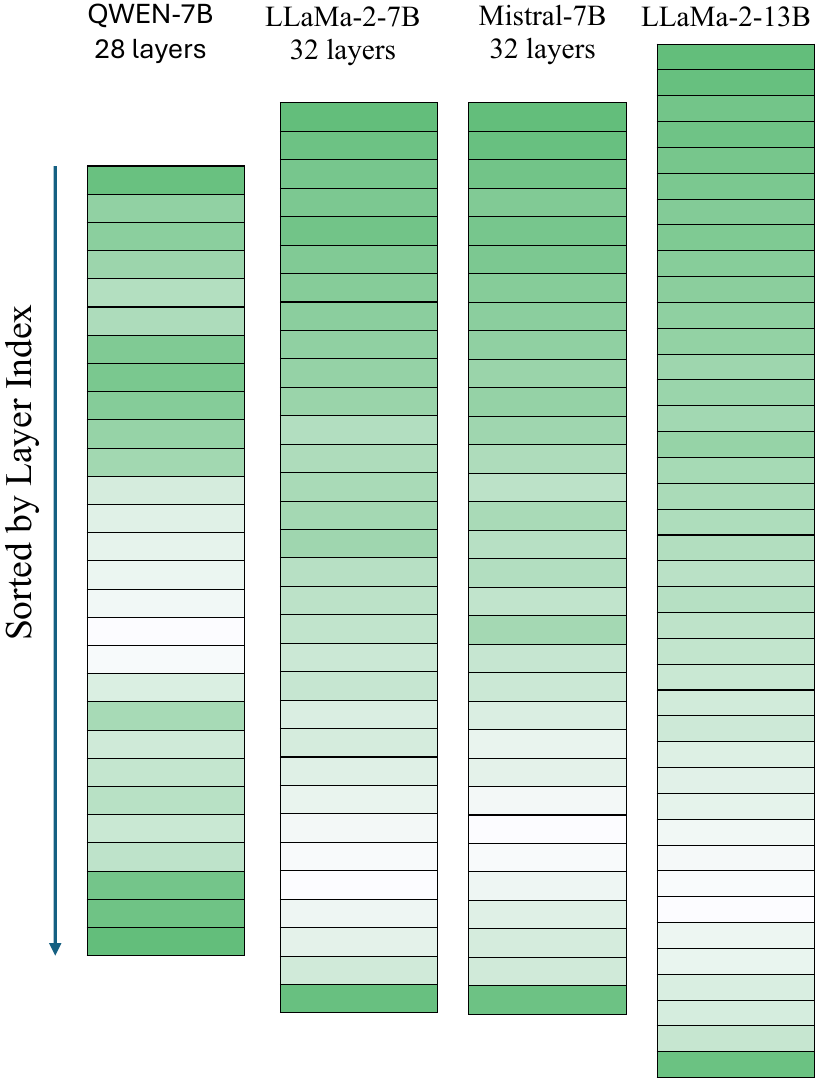} 
\caption{\footnotesize Visualization of the layer importance score for four different LLMs. 
Shown here is our Layer Input Modification (LIM) score.
The color intensity of each layer, which represents their LIM importance score (darker color indicates higher importance score), highlights that the original layer structure does {\em not} have the layers sorted according to their importance.}
\label{LIM_layerrankingCommon}
\end{figure}

\begin{figure*}[h]
\centering
\includegraphics[width=1.0\textwidth]{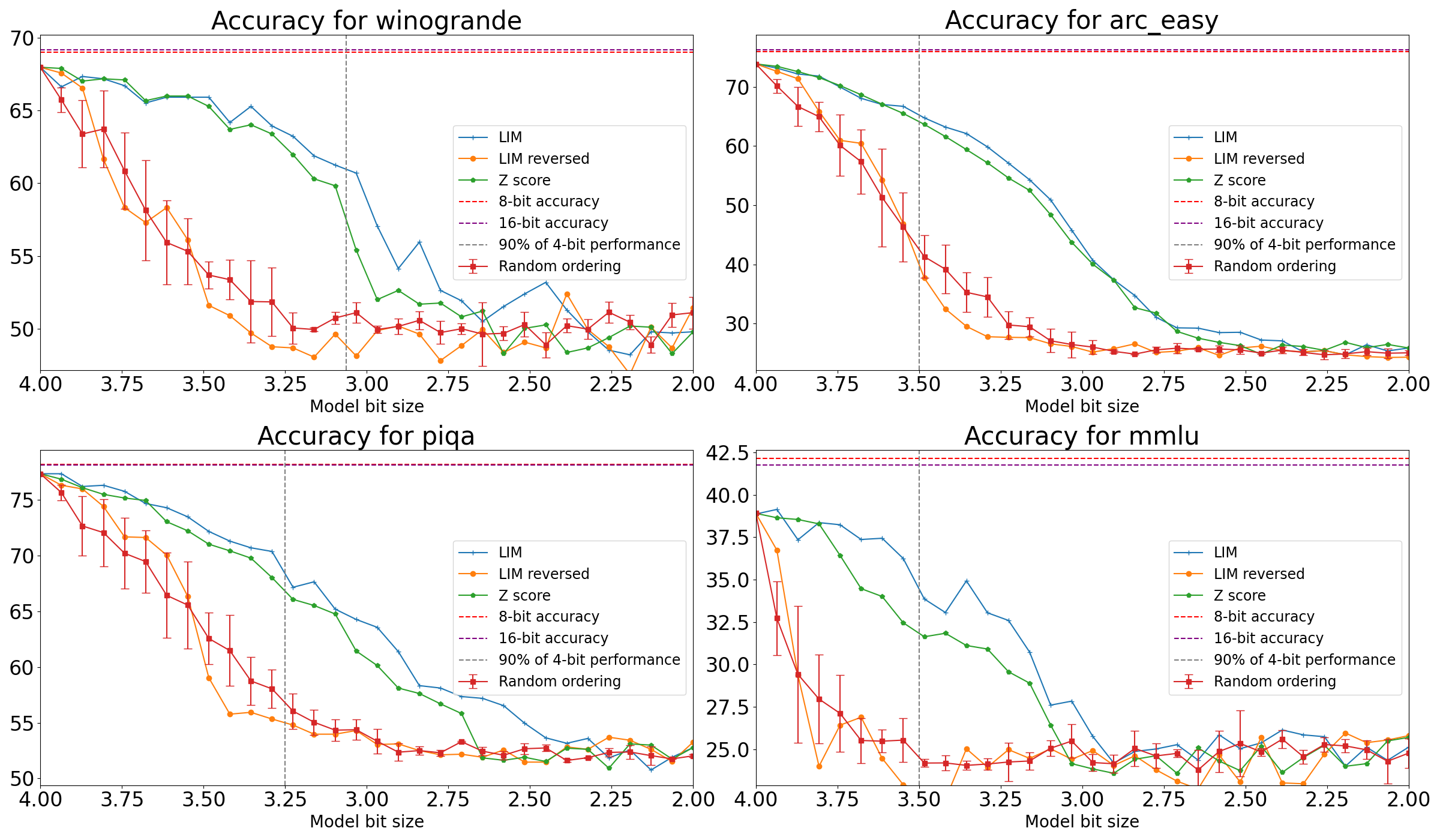}
\caption{Similarly to the other bit plots the graphs showcase the accuracy on four distinct data sets when quantizing LLaMa2-7b from full 4-bits quantization to 2-bit by moving less important layers in 2-bits quantization}
\label{llama7b24bits}
\end{figure*}

\begin{figure*}[h]
\centering
\includegraphics[width=1.0\textwidth]{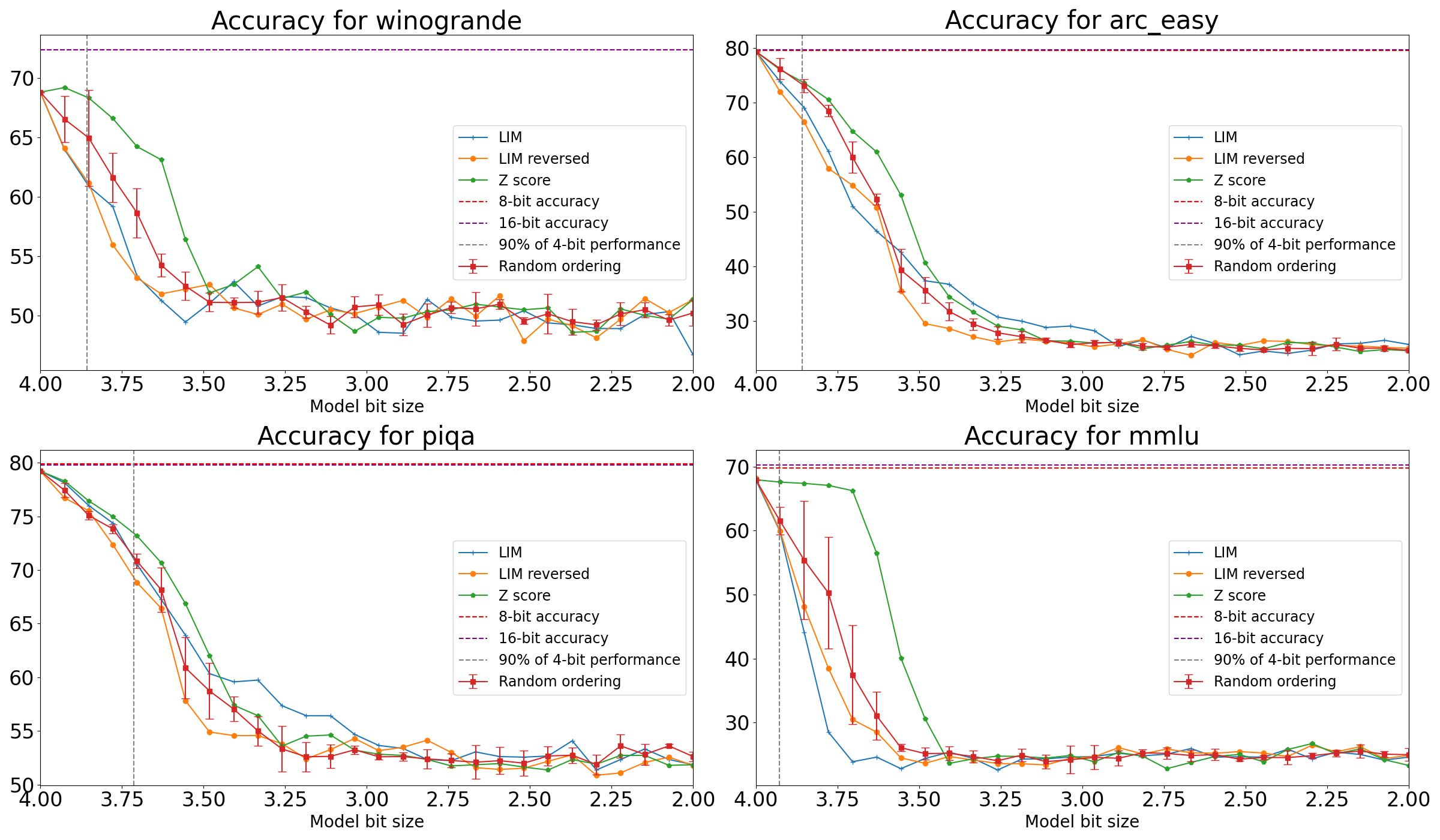}
\caption{Qwen2 quantized with quanto between 4 and 2 bits. All notations are same as in \cref{llama13b24bits}.}
\label{qwen7b24bits}
\end{figure*}

\begin{figure*}[h]
\centering
\includegraphics[width=1.0\textwidth]{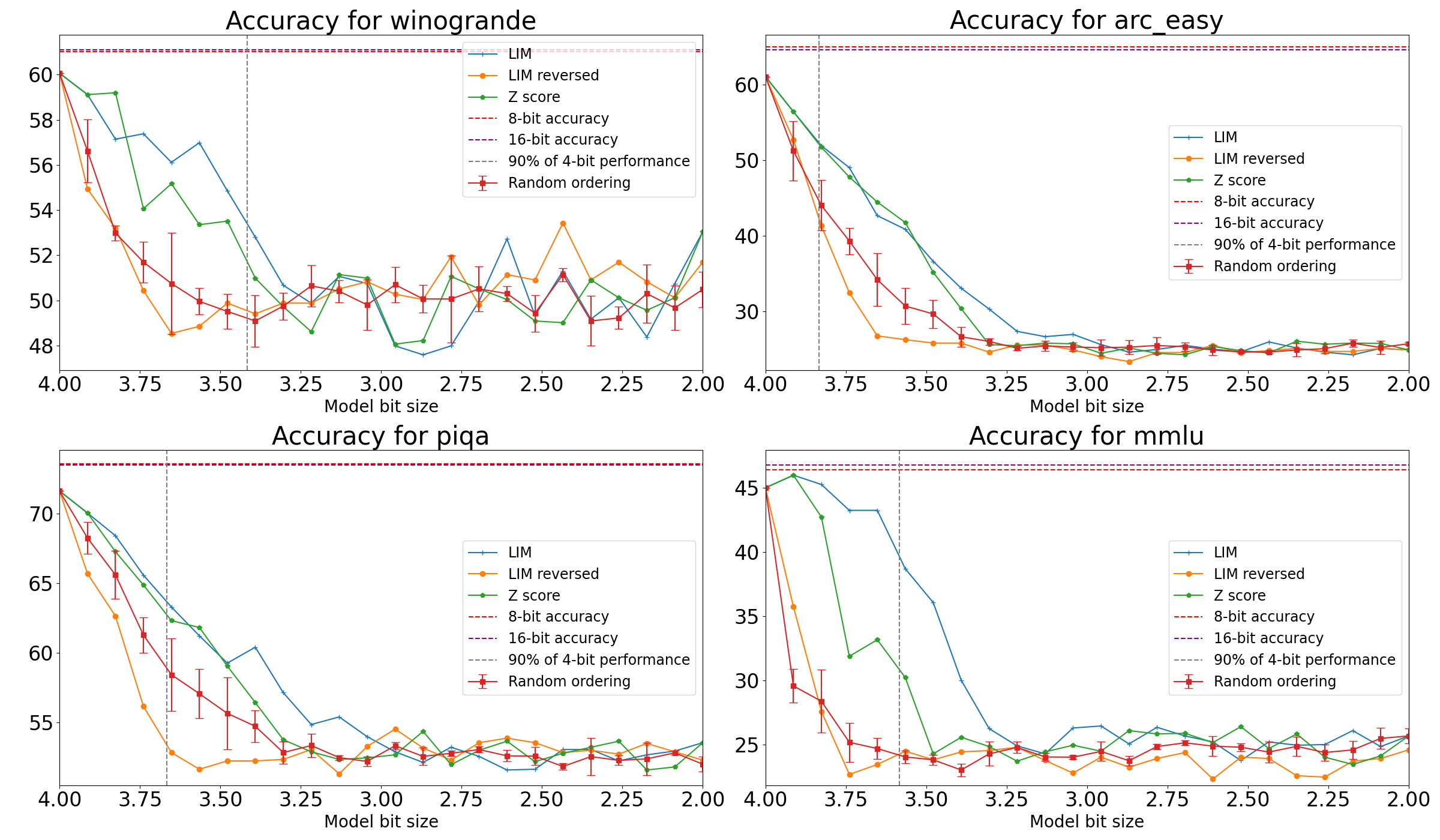}
\caption{Qwen1.5-1.8b quantized with quanto between 4 and 2 bits. All notations are same as in \cref{llama13b24bits}.}
\label{qwen18b24bits}
\end{figure*}

\begin{table*}[ht]    
    \centering    
    \begin{adjustbox}{width=\textwidth}    
    \begin{tabular}{c|c|c|ccccc|c}    
        \toprule    
        & Models & Layers & \textbf{WNGD} & \textbf{ARC} & \textbf{PIQA} & \textbf{HLSWG} & \textbf{MMLU} & \textbf{Average} \\    
        & & low-bits &  & & & & & Accuracy \\    
        \midrule    
        \multirow{12}{*}{\rotatebox[origin=c]{90}{{\bf BI Ordering}}} & \multirow{3}{*}{LLaMa2-7B} & 5 & 69.06 & 75.88 & 77.69 & 57.05 & 41.28 & 64.2 \\     
        & & 10 & 68.82 & 76.3 & 77.42 & 57.11 & 41.96 & 64.3 \\     
        & & 15 & 68.82 & 76.3 & 77.42 & 57.11 & 41.96 & 64.3 \\     
        \cline{2-9}    
        \customstrut{2.5ex}
        & \multirow{3}{*}{Mistral-7B} & 5 & 73.87 & 81.01 & 80.9 & 61.13 & 58.56 & 71.1 \\     
        & & 10 & 73.95 & 80.3 & 80.95 & 61.28 & 58.39 & 71.0 \\     
        & & 15 & 73.79 & 80.42 & 80.79 & 61.17 & 58.14 & 70.9 \\     
        \cline{2-9}    
        \customstrut{2.5ex}
        & \multirow{3}{*}{LLaMa2-13B} & 5 & 72.29 & 79.33 & 79.16 & 60.15 & 50.49 & 68.3 \\     
        & & 10 & 71.74 & 79.08 & 79.32 & 59.96 & 50.53 & 68.1 \\     
        & & 15 & 71.74 & 79.37 & 79.32 & 59.96 & 50.55 & 68.2 \\     
        \cline{2-9}    
        \customstrut{2.5ex}
        & \multirow{3}{*}{Qwen-2-7B} & 5 & 70.71 & 79.2 & 80.03 & 58.66 & 68.06 & 71.3 \\     
        & & 10 & 70.95 & 79.2 & 79.97 & 58.44 & 67.68 & 71.2 \\     
        & & 15 & 68.82 & 78.32 & 79.81 & 58.26 & 67.66 & 70.6 \\     
        \midrule    
        \multirow{12}{*}{\rotatebox[origin=c]{90}{{\bf Z Ordering}}} & \multirow{3}{*}{LLaMa2-7B} & 5 & 68.82 & 76.13 & 77.91 & 57.02 & 40.85 & 64.1 \\     
        & & 10 & 68.66 & 75.92 & 77.63 & 57.09 & 40.6 & 64.0 \\     
        & & 15 & 68.27 & 75.54 & 77.42 & 57.09 & 39.7 & 63.6 \\     
        \cline{2-9}    
        \customstrut{2.5ex}
        & \multirow{3}{*}{Mistral-7B} & 5 & 73.87 & 80.76 & 80.9 & 61.24 & 58.62 & 71.1 \\     
        & & 10 & 74.19 & 80.47 & 81.12 & 61.03 & 58.2 & 71.0 \\     
        & & 15 & 74.42 & 80.47 & 80.84 & 60.89 & 58.31 & 71.0 \\     
        \cline{2-9}    
        \customstrut{2.5ex}
        & \multirow{3}{*}{LLaMa2-13B} & 5 & 72.29 & 79.54 & 78.94 & 60 & 50.54 & 68.3 \\     
        & & 10 & 72.21 & 79.58 & 79.16 & 59.99 & 50.51 & 68.3 \\     
        & & 15 & 72.05 & 79.46 & 79.37 & 59.98 & 50.59 & 68.3 \\     
        \cline{2-9}    
        \customstrut{2.5ex}
        & \multirow{3}{*}{Qwen-2-7B} & 5 & 71.58 & 79.2 & 79.76 & 59.2 & 69.44 & 71.8 \\     
        & & 10 & 71.5 & 79.08 & 80.35 & 58.82 & 69.01 & 71.8 \\     
        & & 15 & 71.58 & 78.61 & 79.92 & 58.54 & 68.65 & 71.5 \\     
        \bottomrule    
    \end{tabular}    
    \end{adjustbox}    
    \caption{Accuracy results of different models across various tasks for 8bit and 4bit quantization using Quanto as the quantization technique.}  
    \label{tab:84bitQuantoResults}    
\end{table*}

\subsection{LLaMa2 and QWEN plots}

We show 4-bits to 2-bits variable layer-wise quantization for LLaMa2-7b and QWEN-7B in \cref{llama7b24bits} and \cref{qwen7b24bits} respectively. All notations are same as in \cref{llama13b24bits}. These curves were also generated on 2K evaluation instances from each of the datasets.

\subsection{Commonality between layer importance}

To the best of our knowledge, we are the first one to propose layer importance and utilize the importance order to quantize different layers at different bits. in \Cref{LIM_layerrankingCommon}, we show the intensity bars below for each layer based on their importance for four different LLMs with different number of layers and sizes. As observed, there is a substantial pattern overlap of least important layers across multiple LLMs. We observe that first and the last layer are the most two important layers. Many of the important layers also tend to be the initial few set of layers. Lesser important layers (shown by block of layers with faded intensity) tend to be towards halfway of middle and end of the network. 
These observations suggest generalized patterns in layer importance across LLMs and pre-computed layer importance orders can be roughly utilized to quantize a wide variety of LLMs.

It is worth noting that (unpublished) concurrent works like \cite{gromov2024unreasonable} have presented an empirical finding that layers towards the end of the model can be removed except the last layer. The reverse order of layers indexes surprisingly has overlap with the importance order calculated with our LIM score but we believe our LIM score is more broadly applicable to any LLM where even the layers towards the end can be more important.


\subsection{Results of 8-bits to 4-bits quantization on different datasets}

We show results of quantizing models lower than 8-bits. Following our proposed methodology, the same techinque can be applied to have the more important layers in 8-bits and the least important ones in 4-bits. While this does still increase the performance that can be fit within a memory requirement, the results are not as major as the ones for the 4-2 bit range, thus we mainly focus on that range.

\end{document}